\documentclass[letterpaper, 10 pt, conference]{ieeeconf}  

\IEEEoverridecommandlockouts
\usepackage[caption=false, font=footnotesize]{subfig}
\usepackage{xcolor}
\usepackage{epsfig} %
\usepackage{graphics} %
\usepackage{times} %
\usepackage{amsmath} %
\usepackage{booktabs, multirow}
\usepackage{soul}%
\usepackage{changepage,threeparttable} %
\usepackage{amssymb}  %
\usepackage{subfig}
\usepackage{graphicx}
\usepackage{lipsum}  
\usepackage{amssymb}
\usepackage{array}
\usepackage{flushend}
\usepackage{cite}
\usepackage{hyperref}
\usepackage{float}
\usepackage{microtype}
\usepackage{xargs}
\usepackage{booktabs}
\usepackage{bm}
\usepackage{dsfont}
\usepackage{tcolorbox}

\usepackage[capitalise]{cleveref}
\usepackage{float}

\title{\tt \LARGE \bf 
ReMoSPLAT: Reactive Mobile Manipulation Control on a Gaussian Splat}
\author{Anonymous Submission}

\author{Nicolás Marticorena, Tobias Fischer, Niko Suenderhauf %
\thanks{The authors are with the QUT Centre For Robotics, School of Electrical Engineering and Robotics at the Queensland University of Technology, Brisbane, QLD 4000, Australia. This research was partially supported by the QUT Centre for Robotics and by funding from ARC DECRA Fellowship DE240100149 to TF. Email: {\tt\small nicolas.marticorena@hdr.qut.edu.au}}}

\usepackage{amsmath}
\usepackage{amssymb}
\usepackage{accents}
\usepackage{bm}
\usepackage{xifthen}
\usepackage{color}
\usepackage{fancyvrb}
\usepackage{graphicx,scalerel}

\newcommand{\ba}{\begin{eqnarray}}
\newcommand{\ea}{\end{eqnarray}}

\newcommand{\presup}[1]{\,{}^{\scriptscriptstyle #1}\!}

\newcommand{\pose}[1][ZZZZ]{\ifthenelse{\equal{#1}{ZZZZ}}{}{\presup{#1}}{\mathbf{\xi}}}
\newcommand{\estpose}[1][ZZZZ]{\ifthenelse{\equal{#1}{ZZZZ}}{}{\presup{#1}}{\mathbf{\hat{\xi}}}}
\newcommand{\hpose}[1][ZZZZ]{\ifthenelse{\equal{#1}{ZZZZ}}{}{\presup{#1}}{\hat{\mathbf{\xi}}}}
\newcommand{\posedot}[1][ZZZZ]{\ifthenelse{\equal{#1}{ZZZZ}}{}{\presup{#1}}{\mathbf{\nu}}}

\newcommand{\q}[1][ZZZZ]{\ifthenelse{\equal{#1}{ZZZZ}}{}{\presup{#1}}{\mathring{q}}}

\DeclareMathAlphabet{\mathitbf}{OML}{cmm}{b}{it}
\newcommand{\twist}[2][ZZZZ]{\ifthenelse{\equal{#1}{ZZZZ}}{}{\presup{#1}}{\mathcal{S}}}
\renewcommand{\vec}[2][ZZZZ]{\ifthenelse{\equal{#1}{ZZZZ}}{}{\presup{#1}}{\mathitbf{#2}}}

\newcommand{\hvec}[2][ZZZZ]{\ifthenelse{\equal{#1}{ZZZZ}}{}{\presup{#1}}{\tilde{\vec{#2}}}}
\newcommand{\obvec}[2][ZZZZ]{\ifthenelse{\equal{#1}{ZZZZ}}{}{\presup{#1}}\rlap{${\overbridge{\phantom{$\vec{#2}$}}}$}\vec{#2}}
\newcommand{\evec}[2][ZZZZ]{\ifthenelse{\equal{#1}{ZZZZ}}{}{\presup{#1}}{\hat{\vec{#2}}}}
\newcommand{\bvec}[2][ZZZZ]{\ifthenelse{\equal{#1}{ZZZZ}}{}{\presup{#1}}{\bar{\vec{#2}}}}

\newcommand{\dvec}[2][ZZZZ]{\ifthenelse{\equal{#1}{ZZZZ}}{}{\presup{#1}}{\dot{\vec{#2}}}}
\newcommand{\ddvec}[2][ZZZZ]{\ifthenelse{\equal{#1}{ZZZZ}}{}{\presup{#1}}{\ddot{\vec{#2}}}}

\newcommand{\mat}[2][ZZZZ]{\ifthenelse{\equal{#1}{ZZZZ}}{}{\presup{#1}\,}{{\boldsymbol #2}}}
\newcommand{\dmat}[2][ZZZZ]{\ifthenelse{\equal{#1}{ZZZZ}}{}{\presup{#1}\,}{{\dot{\boldsymbol #2}}}}
\newcommand{\emat}[2][ZZZZ]{\ifthenelse{\equal{#1}{ZZZZ}}{}{\presup{#1}\,}{\hat{\boldsymbol#2}}}
\newcommand{\matfn}[3][ZZZZ]{\ifthenelse{\equal{#1}{ZZZZ}}{}{\presup{#1}}{{\mat{#2}}\left(#3\right)}}
\newcommand{\Rt}[2][ZZZZ]{\ifthenelse{\equal{#1}{ZZZZ}}{}{\presup{#1}}{{\bf R}\left(#2\right)}}

\newcommand{\point}[2][ZZZZ]{\ifthenelse{\equal{#1}{ZZZZ}}{}{\presup{#1}}{\mathbf{\mathrm{#2}}}}

\newfont{\School}{pncr}
\newfont{\eightTR}{pncr at 8pt}

\usepackage{color}
\usepackage{fancyvrb}
\fvset{formatcom=\color{blue},fontseries=c,fontfamily=courier,xleftmargin=4mm,commentchar=!}

\DefineVerbatimEnvironment{Code}{Verbatim}{formatcom=\color{blue},fontseries=c,fontfamily=courier,fontsize=\footnotesize,xleftmargin=4mm,commentchar=!}

\DefineVerbatimEnvironment{CodeSmall}{Verbatim}{formatcom=\color{blue},fontseries=c,fontfamily=courier,fontsize=\scriptsize,xleftmargin=1mm,commentchar=!}

\DefineVerbatimEnvironment{CodeNum}{Verbatim}{numbers=left,numbersep=4pt,formatcom=\color{blue},fontseries=c,fontfamily=courier,fontsize=\footnotesize,xleftmargin=4mm}

\newcommand{\model}[1]{\index{code}{#1@\textit{#1}}\ifthenelse{\boolean{draft}}{{\color{green}\Verb+#1+}}{\Verb+#1+}}
\newcommand{\block}[1]{\ifthenelse{\boolean{draft}}{{\color{green}\Verb+#1+}}{\textsf{#1}}}

\newcommand{\func}[2][ZZZZ]{\ifthenelse{\equal{#1}{ZZZZ}}{\index{code}{#2}}{\index{code}{#1}}\ifthenelse{\boolean{draft}}{{\color{green}\Verb+#2+}}{\Verb+#2+}}

\newcommand{\methodb}[2]{\index{code}{#1@\textbf{#1}!.#2}\ifthenelse{\boolean{draft}}{{\color{magenta}\Verb+#1.#2+}}{\Verb+#1.#2+}}
\newcommand{\method}[2]{\index{code}{#1@\textbf{#1}!.#2}\ifthenelse{\boolean{draft}}{{\color{magenta}\Verb+#2+}}{\Verb+#2+}}
\newcommand{\class}[1]{\index{code}{#1@\textbf{#1}}\ifthenelse{\boolean{draft}}{{\color{cyan}\Verb+#1+}}{\Verb+#1+}}
\newcommand{\property}[1]{\index{property}{#1}\ifthenelse{\boolean{draft}}{{\color{cyan}\Verb+#1+}}{\Verb+#1+}}

\newcommand{\SE}[1]{\ensuremath{\mathrm{{\bf SE}(#1)}}}
\newcommand{\SO}[1]{\ensuremath{\mathrm{{\bf SO}(#1)}}}

\begin{document}

\bstctlcite{bstctl:forced_etal,bstctl:nodash}
\maketitle

\begin{abstract}
Reactive control can gracefully coordinate the motion of the base and the arm of a mobile manipulator. However, incorporating an accurate representation of the environment to avoid obstacles without involving costly planning remains a challenge. 
In this work, we present ReMoSPLAT, a reactive controller based on a quadratic program formulation for mobile manipulation that leverages a Gaussian Splat representation for collision avoidance. 
By integrating additional constraints and costs into the optimisation formulation, a mobile manipulator platform can reach its intended end effector pose while avoiding obstacles, even in cluttered scenes. We investigate the trade-offs of two methods for efficiently calculating robot-obstacle distances, comparing a purely geometric approach with a rasterisation-based approach.
Our simulation experiments on both synthetic and real-world scans demonstrate the feasibility of the proposed method, achieving performance comparable to controllers that rely on perfect ground-truth information. We further validate the approach on a real robot platform more details:
\url{https://remosplat.github.io}
\end{abstract}

\section{Introduction}
Reactive control is an effective approach to control a mobile manipulator towards a desired target pose, without requiring costly global planning~\cite{Haviland2022, Heins2021, local_reactive}. However, how the robot should perceive and represent the geometry of its surroundings to obtain collision-free motion in such a scenario is still an open challenge~\cite{Sereinig2020}. This problem is often overlooked~\cite{Haviland2022}, oversimplified~\cite{Heins2021}, or assumed to be solved~\cite{local_reactive}.

Traditionally, scene representations for control and planning have relied on occupancy grids~\cite{Arbanas2018DecentralizedTeams}, meshes~\cite{rgb}, or point clouds~\cite{Tulbure2020}.
More recently, Gaussian Splatting (GS)~\cite{Kerbl20233DRendering} has emerged as an explicit scene representation that is efficient to render and represents the underlying scene geometry with sufficient fidelity to support downstream tasks in robotics~\cite{Zhu20243DSurvey}.
For example, recent works have explored leveraging the geometric information encoded in GS representations for robotic navigation~\cite{Chen2024Splat-Nav:Maps, Andreu2025FOCI:Splats,  ChenSAFER-SplatMaps}, where each Gaussian is interpreted as an obstacle in the environment. However, the potential of Gaussian Splatting as a scene representation for reactive mobile manipulation control remains unexplored.

\begin{figure}[t]
    \centering
    \includegraphics[width=0.9\linewidth]{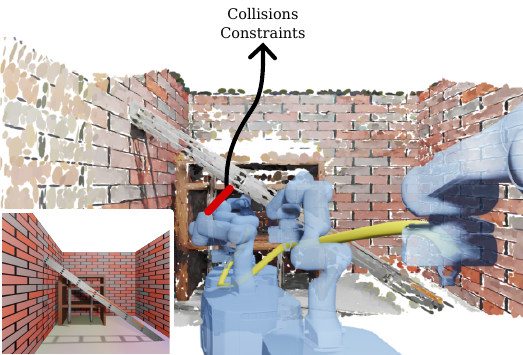}
    
    \vspace*{-0.2cm}
    \caption{
    Illustration of our reactive control approach for a mobile manipulator tasked with reaching a 3D pose inside a bookshelf.
    The scene is represented as a 2D Gaussian Splat, visualised as ellipsoids in the Front view.
    The controller computes joint velocities to drive the robot toward the target pose, producing the yellow end-effector and blue base trajectories. Collision avoidance is enforced through hard and soft constraints, shown as red line segments derived from distance estimates against the Gaussian Splat.
    }
        \label{fig:main_figure}
        \vspace*{-0.3cm}
\end{figure}

In this work, we explore bridging modern differential rendering based GS scene representations with a established control framework for mobile manipulation, enabling for the first time collision-aware mobile manipulation directly over a GS. As illustrated in~\cref{fig:main_figure}, we show how collision avoidance can be integrated into a reactive controller by incorporating additional constraints and cost terms into a Quadratic Program (QP). Importantly, we show how both the geometry (i.e., position and shape) \emph{and} the opacity of the Gaussian Splats must be considered for successful obstacle avoidance in cluttered real scenes. 
Our contributions are:
\begin{itemize}
    \item An extension of a QP-based reactive mobile manipulation controller that enables collision avoidance directly on a Gaussian Splat scene representation.
    \item An investigation of the trade-offs in accuracy and efficiency arising from two strategies to calculate robot-obstacle distances: a purely geometric approach and a rasterisation-based method that accounts for opacity.
\end{itemize}

\section{Related Works}

\subsection{Collision avoidance in motion generation}
Collision avoidance is a fundamental aspect of motion generation.
Approaches range from modelling repulsive forces, as in artificial potential fields~\cite{Tulbure2020}, to collision checking routines~\cite{Schulman2014MotionChecking}, and as constraints or collision costs in optimisation-based approaches~\cite{Zucker2013}. A common factor across these methods is that their quality depends on the underlying scene representation.

In structured environments, the scene can be represented by either geometric primitives or convex shapes~\cite{Tracy2023DifferentiablePrimitives}. However, when operating in unstructured or unseen scenes, the robot must rely on a representation built from sensor information.
Examples of popular representations include occupancy maps such as \textit{OctoMap}~\cite{Hornung2013}, which represent space as either “occupied” or “free”; Signed Distance Fields (SDFs)~\cite{Pan2022Voxfield:Reconstruction}, which provide direct distance information and can be incorporated into optimisation-based planners~\cite{Zucker2013} or controllers~\cite{Pankert2020}; and, in some cases, raw point clouds obtained from external sensors~\cite{Tulbure2020}.

Building on the idea of utilising a 3D representation for collision avoidance, we leverage Gaussian Splatting, which provides an explicit, sensor-derived representation that is both rich in geometric information and efficient to render.

\subsection{Gaussian Splatting in robotics}

Gaussian Splatting (GS)~\cite{Kerbl20233DRendering} has been of great interest in the robotics community, due to its high-quality rendering capabilities and significantly faster rendering compared to previous differential rendering approaches such as NeRFs~\cite{Mildenhall2020}.
GS has been applied within the robotics field to various tasks, including visual SLAM~\cite{keetha2024splatam,Matsuki:Murai:etal:CVPR2024,sandstrom2024splat}, grasp proposal~\cite{Ji2024GraspSplats:Splatting, GaussianGrasper} and physics simulation~\cite{xie2024physgaussian}.
For a comprehensive survey of GS in robotics, we refer the reader to~\cite{Zhu20243DSurvey}.

Here, we focus on works that employ GS for motion generation. 
The works of Splat-Nav~\cite{Chen2024Splat-Nav:Maps} and FOCI~\cite{Andreu2025FOCI:Splats} focus on robot navigation planning.
In both approaches, the robot is represented by a set of Gaussians, and the approaches differ by either using an intersection check to compute safe polytope corridors~\cite{Chen2024Splat-Nav:Maps}, or directly using the overlap as a collision cost~\cite{Andreu2025FOCI:Splats}. 
The work of Safer-Splats~\cite{ChenSAFER-SplatMaps} introduced a control barrier function (CBF) to update velocity commands to keep a collision-free motion. Finally, the work of Splanning~\cite{splanning} introduced a trajectory optimisation for serial robot manipulators in which a reformulated normalised Gaussian Splat is used to bound a collision probability.

In contrast, our work introduces a QP-based controller for mobile manipulation that extracts distances to surfaces encoded in the splats. We further investigate alternative formulations for distance extraction, including methods that incorporate Gaussian opacity.

\section{Background}
We briefly summarise the controller we built upon (\cref{sec:Controller}) and the Gaussian Splatting (GS) representation used in this work (\cref{sec:3DGS}), with particular focus on geometry-accurate variants (\cref{sec:depth}).

\subsection{Mobile manipulation reactive controller}\label{sec:Controller}

Our approach is based on the holistic mobile manipulation controller presented in~\cite{Haviland2022}, formulated as a quadratic program (QP) as:
\begin{align}
    \min_{\vec{x}} \frac{1}{2} \vec{x}^\top \mat{Q} \vec{x} &+ \vec{c}^\top \vec{x}\label{eq:qp_inf}\\
    \text{s.t} \text{  }\mathcal{J}\vec{x} &= \vec[b]{v}_e \label{eq:equality}\\
    \mat{A} \vec{x} &\leq \vec{b} \label{eq:ab}\\
    \chi^{-} &\leq \vec{x} \leq \chi^{+}\label{eq:limits}.
\end{align}
Here the decision variable $\vec{x}$ is a combination of the joint velocities $\dvec{q}$ and a slack component $\delta \in \mathbb{R}^6$:
\begin{equation}
    \vec{x} = \begin{pmatrix}\dvec{q}\\ \vec{\delta}\end{pmatrix} 
\in \mathbb{R}^{(n_{\text{dof}} + 6)},
\end{equation}
where $n_{\text{dof}} = |\vec{q}|$ is the degrees of freedom of the robot.

The quadratic cost function $\mat{Q}$ defined in \cref{eq:qp_inf} is designed to minimise joint velocities; the linear cost component $\vec{c}$ is given by:
\begin{equation}\label{eq:qp_c}
    \vec{c} = \begin{pmatrix}
        \mat{J}_\text{m} +  \mat{J}_\text{o} \\ \mathbf{0}_{6}
    \end{pmatrix} \in \mathbb{R}^{(n_{\text{dof}}+6)},
\end{equation}
with $\mat{J}_\text{m} \in \mathbb{R}^{n_{\text{dof}}}$ representing the manipulability Jacobian~\cite{Haviland2020a} that maximises the manipulability index of the robot arm~\cite{Yoshikawa1985}. The term $\mat{J}_\text{o} \in \mathbb{R}^{n_{\text{dof}}}$ comprises a linear component to optimise the base's relative orientation to the end effector~\cite{Haviland2022}.

To address collision avoidance, we followed a similar methodology introduced in~\cite{Marticorena2024RMMI:Map}, but completely removing its reliance on an SDF in favor of using a GS representation.

\subsection{3D Gaussian Splatting}\label{sec:3DGS}
Gaussian Splatting, first introduced in~\cite{Kerbl20233DRendering}, is a technique for explicitly reconstructing a scene using a set of anisotropic 3D Gaussians, enabling real-time high-quality rendering from novel viewpoints. Each reconstructed scene is represented by a set of Gaussians,
%
where each Gaussian is parametrised by a set of trainable parameters, including its position $\mu_i\in\mathbb{R}^3$, 3D covariance $\Sigma_i \in \mathbb{R}^{3\times3}$, opacity $o_i \in [0,1]$ and colour $c_i$ represented as coefficients of spherical harmonics to map view-dependent appearance.

Training begins by initialising the Gaussians from a sparse point cloud reconstructed via Structure-from-Motion (SfM)~\cite{schoenberger2016sfm}. 
The scene can then be rendered through differential rasterisation, consisting of first projecting (splatting) the 3D Gaussians in the image plane.
Each pixel in the image is then obtained by $\alpha$-blending with the Gaussians sorted in depth order. The parameters of each Gaussian are then updated using stochastic gradient descent with a photometric loss between the rendered images and captured images.

As our work utilises GS to model geometry, we focus our attention on GS variations that prioritise geometry-accurate reconstructions, which we discuss in the following.

\subsection{Geometry-accurate Gaussian Splatting}\label{sec:depth}
Gaussian Splatting is an active area of research, with algorithmic variations emerging across different applications. One key sub-field focuses on surface reconstruction, also known as geometry-accurate methods, which aim to improve both rendering fidelity and reconstruction quality~\cite{Huang2024,Zhang2024,Guedon2024,Wei2024,Yu2024}.

These methods generally fall into two categories: dual-optimisation approaches, where a surface model is learned jointly with the Gaussian Splats~\cite{Wei2024,Yu2024}, and depth rasterisation methods, which enhance geometry extraction from rendered views~\cite{Huang2024,Zhang2024,Guedon2024}.

Our work builds on the latter, using 2D Gaussian Splat~\cite{Huang2024} due to the availability of an open-source implementation during development of our method\footnote{\url{https://github.com/nerfstudio-project/gsplat}}. However, any depth rasterisation method could be used, meaning advances in this area could directly benefit our approach.

For rasterising depth, the median depth approach proposed in~\cite{Huang2024} is used. 
This method computes the depth of a ray as:
\begin{equation}\label{eq:depth_median}
    z_{\text{median}} = \max \left\{z_i \mid T_i > 0.5\right\},
\end{equation}
where $z_i$ is the distance along the camera's $z$-axis at which a ray intersects the $i$-th 2D ellipsoid, sorted by depth. The term $T_i\in[0,1]$ indicates the transmittance, i.e. the probability that the ray passes through all splats up to index $i$, computed as:
\begin{equation}
    T_i = \prod_{k=1}^{i-1}(1-o_k).
\end{equation}
\section{Methodology}
Our method, ReMoSplat, focuses on controlling a mobile manipulator to a desired target end effector pose using a purely kinematic reactive approach (\cref{sec:system}).
At each control step, the controller uses the robot’s current configuration together with a Gaussian Splat (GS) representation that reflects the current estimate of the scene geometry.
To obtain collision-free motions, we require a function of the robot's distance from the underlying surface, in which we simplify the robot's geometry by representing it as a set of spheres (\cref{sec:geometry}). \cref{sec:distance} explores two approaches to obtain distances to the closest surface in a GS representation: (i) a sphere-to-ellipsoid computation and (ii) a rasterisation-based method using \textit{virtual cameras} (see also \cref{fig:DiagramDistance}). Finally, we show how to employ this information within a QP-based reactive manipulation controller, both as inequality constraints (\cref{sec:collision_constrains}) and an active avoidance collision cost (\cref{sec:active_collision}).


    

\subsection{System modelling}\label{sec:system}
Similar to~\cite{Haviland2022}, we adopt a holistic kinematic model that enables coordinated control of the mobile base and the manipulator arm. 
The robot state is defined by the 2D base pose $(x_b, y_b, \theta_b)$ and the arm joint configuration $\vec{q}_{\text{arm}}$. 
The end effector pose in the world frame, $\mat[w]{T}_{e} \in \SE{3}$ can be obtained by forward kinematics as:
\begin{equation}
    \mat[w]{T}_{e} = \mat[w]{T}_b(x_b,y_b,\theta_b) \cdot \mat[b]{T}_a \cdot \mat[a]{T}_e(\vec{q}_{\text{arm}}),
\end{equation}
where $\mat[w]{T}_b(x_b,y_b,\theta_{b})\in\SE{3}$ is the base pose in the world frame, 
$\mat[b]{T}_a \in \SE{3}$ is the constant transformation from the base to the arm mount, 
and $\mat[a]{T}_e(\vec{q}_{\text{arm}}) \in \SE{3}$ is the forward kinematics of the arm.

The degrees of freedom of the base are modelled as two virtual joints $\delta_x$ and $\delta_{\theta}$ as the infinitesimal forward translation and rotation. We can represent the controllable joints of the base as $\vec{q}_{\text{base}} = (\delta_x, \delta_{\theta})$, and the overall joint state of the robot as $\vec{q} = (\vec{q}_{\text{base}}, \vec{q}_{\text{arm}})$.

Finally, we can obtain the differential kinematic map of the end effector velocity measured in the world frame $\vec[w]{v}_e \in \mathbb{R}^6$, with respect to the robot joint velocity $\dvec{q} = (\dvec{q}_\text{base},\dvec{q}_\text{arm})\in \mathbb{R}^{n_\text{dof}}$
\begin{equation}
    \vec[w]{v}_{e} = \mat[w]{J}_e(x_b,y_b,\theta_b,\vec{q})\begin{pmatrix} \dvec{q}_{\text{base}}\\ \dvec{q}_{\text{arm}} \end{pmatrix},
\end{equation}
where $\mat[w]{J}_e\in\mathbb{R}^{6\times n_{\text{dof}}}$ is the manipulator Jacobian.

\subsection{Robot geometry approximation}\label{sec:geometry}
As is standard practice, we approximate the robot geometry using a set of $N$ spheres~\cite{Zucker2013, Sundaralingam2023CuRobo:Generation, 2025foamtoolsphericalapproximation}.
 Each sphere is characterised by a radius $r \in \mathbb{R}^+$, a position ${}^{\ell}\vec{p} \in \mathbb{R}^3$ relative to a specific robot's link $\ell$ and the pose of the robot's $\ell$-th link $\mat{T}_{\ell} \in \SE{3}$.
 We represent the full set of spheres as:
 \begin{equation}
    \mathcal{R} = \left\{ (r_j, {}^{\ell_j}\vec{p_j}, \mat{T}_{\ell_j})\right\}_{j=1}^{N},
\end{equation}
where $\ell_j$ denotes the index of the robot link to which the $j$-th sphere is attached.
%

\subsection{Distance query}\label{sec:distance}

While Gaussian Splats provides an explicit representation of a scene as discussed in \cref{sec:3DGS}, computing distances from the robot to the environment is not as trivial as it may first appear. One might assume it suffices to measure the distance from each robot sphere to the closest ellipsoid, but this overlooks the opacity information stored in each of the Gaussians. We evaluate two complementary distance approaches. The first is a direct sphere-to-ellipsoid geometric distance computation (\cref{sec:euclidean}), which captures individual splat geometry but ignores opacity. The second is a multi-view depth rasterisation approach (\cref{sec:distance_depth}), which aggregates contributions through rasterisation and accounts for opacity.
As our experiments in \cref{sec:exp3} will reveal, the direct sphere-to-ellipsoid computation, while perhaps having a simpler implementation, is fragile in scenes with low-opacity ellipsoids, whereas the approach that relies on depth rasterisation still holds in such scenes.

\subsubsection{Sphere-to-ellipsoid}\label{sec:euclidean}
The first approach consists of computing the minimum distance between a sphere, whose centre is located at $\vec{p}_r\in\mathbb{R}^3$, and the surface of an ellipsoid.
To do so, we first find the closest point $\vec{p}_e\in \mathbb{R}^3$ on the ellipsoid surface.
The ellipsoid is defined as the level set of a 3D quadratic form (i.e., a Gaussian at a chosen confidence scale), and the distance can be formulated as the following constrained optimisation problem:

\begin{figure}[t]
\centering
\includegraphics[width=0.9\linewidth]{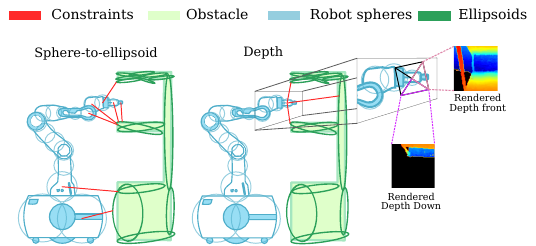}
\centering
\vspace{-8pt}
\caption{High-level diagram of our approach, showing the same robot state and how it models the constraints depending on the chosen distance method. (Left) Illustrates how the sphere-to-ellipsoid generates constraints based on the closest point on the ellipsoid surface of each sphere (\cref{sec:euclidean}). (Right) Illustrates how the depth rasterisation approach relies on rendering depth images from candidate orientations (\cref{eq:rotations}), resulting in one collision constraint per rendered view (\cref{sec:distance_depth}).}
\label{fig:DiagramDistance}
\end{figure}

\begin{equation}
    \begin{aligned}
        &\min_{\vec{p}_e} \quad \left\| \vec{p}_r - \vec{p}_e \right\|_{2}
        &\text{subject to} \quad \vec{p}_{e}^{\top} \mat{C} \vec{p}_{e} = 1,
    \end{aligned}
\end{equation}
where $\mat{C}\in\mathbb{R}^{3\times3}$ is the conic coefficient matrix of the ellipsoid.

This optimisation does not have a known closed-form solution~\cite{Uteshev2018Point-to-ellipseAnalysis}. However, by employing a change of variables, this problem can be reformulated as a convex root-finding problem, which can be solved robustly numerically using the bisection method as shown in~\cite{DistancePointEllipse}.

Once the closest point on the ellipsoid surface is determined, the distance from the $j$-th sphere to this point can be computed as:
\begin{equation}
    d_j =  \left\| \vec{p}_{j} - \vec{p}_e\right\| _2 - r_{j},
\end{equation}
and its gradient:
\begin{equation}
    \nabla_{\vec{p_j}}d_{j} = \frac{\vec{p}_j - \vec{p}_e}{\left\| \vec{p}_{j} - \vec{p}_e\right\| _2},
\end{equation}
where $r_j$ and $\vec{p}_j$ denote the radius and the position of the $j$-th sphere in the robot's spherical approximation (\cref{sec:geometry}).

\subsubsection{Depth rasterisation}\label{sec:distance_depth}
The sphere-to-ellipsoid method ignores opacity, treating all ellipsoids as fully occupied. As a complementary strategy, we build on existing GS depth rasterisation methods \cite{Huang2024}. While the rendering itself follows these prior approaches, we introduce a virtual distance sensor that aggregates depth information from six \textit{virtual cameras} per sample oriented in different directions.

In order to obtain the distance of the robot to the environment, we place one such virtual sensor on each of the robot's spheres (\cref{sec:geometry}):
\begin{equation}
    \mathcal{D} = \left\{D_j =   (\mat{T}_{D_{j}},r_j, \mat{K}) \;\middle|\;  (r_j,\vec{p}_j,\mat{T}_{j})\in \mathcal{R}  \right\},
\end{equation}
where each sensor $D_j$ is parametrised by their 6-DoF pose $\mat{T}_{D_{j}}\in \SE{3}$, sensor radius $r_j \in {\mathbb{R}}^{+}$ and the virtual camera intrinsics $\mat{K}\in \mathbb{R}^{3\times 3}$ which is shared across all cameras to allow batch rendering.
We set $\mat{K}$ to consider both a horizontal and vertical field of view of $90^{\circ}$ to maximise coverage, accounting for distortions.

Each $j$-th sensor estimates distances by rasterising depth images from six different viewpoints. Each of these six viewpoints is defined by a local rotation with respect to $\mat{T}_{D_{j}}$.
Specifically, to obtain depth images in all directions, the orientation of each $c$-th camera is chosen from a discrete set of candidate rotations $\mathcal{C}_R\subset \SO{3}$ :
\begin{equation}\label{eq:rotations}
    \mathcal{C}_R =  \big\{\mat{R}_x(0),\; \mat{R}_x(\pm\tfrac{\pi}{2}),\; \mat{R}_y(\tfrac{\pi}{2}),\; \mat{R}_y(\pi),\; \mat{R}_y(\tfrac{3\pi}{2}) \big\}. 
\end{equation}
where $\mat{R_x}, \mat{R_y} \in \SO{3}$ are rotations about the $x$- and $y$-axes, respectively.

The pose of each $c$-th camera of the $j$-th distance sensor in the world frame $\mat{T}_c \in \SE{3}$ is obtained by composing the sensor pose $\mat{T}_{D_j}$ with the relative rotation $^{D_j}\mat{R}_c\in\mathcal{C}_R$.
Using these poses and $\mat{K}$, we rasterise depth from the splats following~\cite{Huang2024}, as outlined in \cref{sec:depth}.

The depth rasterisation process returns a set of 6 depth maps $Z^{j}_c[u,v]\in\mathbb{R}^{+}$ for each $j$-th sensor, where each value encodes the distance along the camera's optical ($z$-axis) direction.
Each depth map is backprojected into a 3D point cloud expressed in the $c$-th camera frame:
\begin{equation}
   ^{c}\mathcal{P}_{j} = \left\{ \; ^c\vec{p}(u,v) \;=\; Z^{j}_c(u,v) \, \mat{K}^{-1} \begin{bmatrix} u \\ v \\ 1 \end{bmatrix} \;\middle|\; (u,v) \in \mathcal{I}_c \right\}, 
\end{equation}
where $\mathcal{I}_c$ denotes the pixel domain of camera $c$.
Each point $^c\vec{p}\in\mathcal{P}^j_c$ corresponds to the backprojected 3D location of a pixel in the depth map.

The closest distance of the $j$-th sensor in the $c$-th camera's direction is given by:
\begin{align}
 d_j^{c} = \min_{\vec{p} \in ^{c}\mathcal{P}_j} \| \vec{p} \| - r_j,
\end{align}
with the direction of increase defined by:
 \begin{align}
     \nabla{d}^{c}_j = \mat{K}^{-1}(u_{\min}, v_{\min},1)^\top,
 \end{align}
 where $(u_{\min},v_{\min})$ is the pixel of the closest depth point.

\subsection{Incorporating distances into the controller}\label{sec:collision_constrains}
As discussed in Section~\ref{sec:distance}, all considered distance estimation approaches yield a common output structure: the distance from each $j$-th robot sphere centre to the closest obstacle, denoted as $d_j$, and a corresponding unit vector $\nabla d_j$ that points in the direction of the closest obstacle surface.
For the depth rasterisation approach, each sphere yields six such distances and unit vectors, and for ease of notation, we will drop the $c$ index.
\Cref{fig:DiagramDistance} illustrates both distance methods and how the distance information is integrated.

To incorporate this information into the QP controller (Section~\ref{sec:Controller}), we formulate constraints that regulate the distance between the robot and obstacles. This requires expressing the time derivative of the distance $d_j$ with respect to the control input, i.e., the joint velocities.

Since the distance is a function of the sphere's position, and each sphere position $\vec{p}_j$ is in turn a function of the robot configuration $\vec{q}$ considering only the joints up to the link where the sphere is locally defined, we denote these joints as $\vec{\tilde{q}}\in\mathbb{R}^{k}$, where $k\leq |\vec{q}|$, the time derivative of the distance can be written using the chain rule:
\begin{equation}
    \dot{d}_j = \nabla d_j^\top \dvec{p}_j = \nabla d_j^\top \mat{J}_v(\tilde{\vec{q}}_j)\dot{\tilde{\vec{q}}},
\end{equation}
where $\mat{J}_v(\tilde{\vec{q}}_j) \in \mathbb{R}^{3\times k}$ is the translational part of the manipulator Jacobian at the $j$-th sphere's centre.

We enforce a safety constraint using a velocity damper formulation, which ensures that the rate of approach toward obstacles decreases as the robot gets closer:
\begin{equation}
\mat{J}_d(\tilde{\vec{q}}_j)\dot{\tilde{\vec{q}}}_j \leq \frac{d_j - d_s}{d_i - d_s},
\end{equation}
where $\mat{J}_d(\tilde{\vec{q}}_j) = \nabla d_j^\top \mat{J}_v(\tilde{\vec{q}}_j)  \in \mathbb{R}^{k}$,
$d_i$ is the influence distance (beyond which obstacles are ignored), and $d_s$ is the stopping distance (the minimum acceptable distance to the obstacle). 

Each sphere within the influence distance generates one constraint using the sphere-to-ellipsoid method, and up to six constraints using the depth approach (one per direction). Stacking all such constraints results in an inequality matrix:
\begin{equation}\label{eq:cha4:ine_matrix}
    \underbrace{
    \begin{pmatrix}
        \mat{J}_{d_1}(\tilde{\vec{q}}_1) & \mathbf{0}_{1 \times 6 + n_{\text{dof}}-k_1}\\
        \vdots & \vdots\\
        \mat{J}_{d_N}(\tilde{\vec{q}}_N) & \mathbf{0}_{1 \times 6 + n_{\text{dof}}-k_N}\\
    \end{pmatrix}
    }_{\displaystyle \mat{A}_c}
    \vec{x}
    \leq
    \underbrace{
    \begin{pmatrix}
        \frac{d_1- d_s}{d_i - d_s} \\
        \vdots \\
        \frac{d_N - d_s}{d_i - d_s}
    \end{pmatrix}
    }_{\displaystyle \vec{b}_c },
\end{equation}
where $\vec{x}(t)$ is the QP decision variable. This linear inequality constraint is appended to the main controller formulation by stacking it in the matrix $\mat{A}$ and the vector $\vec{b}$ in \cref{eq:ab}.

\subsection{Active collision avoidance}\label{sec:active_collision}
In addition to the inequality constraints discussed in \cref{sec:collision_constrains}, we utilise a distance cost on the overall optimisation described in~\cite{Marticorena2024RMMI:Map}. This term penalises motions that bring the robot closer to nearby obstacles, thereby encouraging safer motions.

This linear cost $\mat{J}_{\text{collision}} \in \mathbb{R}^{|\vec{x}|}$ is computed as a weighted average of the distance Jacobians $\mat{J}_d$:
\begin{equation}
    \mat{J}_{\text{collision}} = \frac{\sum_{j \in \mathcal{C}} \mat{J}_{d_{j}} w_j}{\sum_{j\in \mathcal{C}} w_j} ,
\end{equation}
where $\mathcal{C}$ denotes the set of active constraints (e.g. spheres within the influence distance), and $w_j$ is a weight inversely proportional to the distance $d_j$, ensuring a higher penalty for closer spheres
\begin{equation}
    w_j = \frac{d_i- d_j}{d_i - d_s}.
\end{equation}

This cost is linear in $\vec{x}$ and is added to the QP objective via the vector $\vec{c}$ in \cref{eq:qp_c}. We use the dynamic gain of~\cite{Marticorena2024RMMI:Map}. Which in practice, enables the robot to proactively choose directions that increase clearance from obstacles, thereby improving safety.

It is important to note that the construction of this cost term does not require querying more information from the scene representation.
This is achieved by an exploitation of the structure of the matrix $\mat{A}_c$ and vector $\vec{b}_c$ presented in \cref{eq:cha4:ine_matrix} that were obtained when formulating the hard constraints, where the components of $\mat{A}_c$ are averaged with a weight based on the distances stored in $\vec{b}_c$.
\section{Evaluation}
We evaluate our method in a set of experiments. In \cref{sec:setup}, we  detail our experimental setup, including the training process of the Gaussian Splats, the robot platform and hardware used and simulation details. \cref{sec:exp1} explains the main experiment, a reaching task over a set of randomised synthetic scenes.
Subsequently, we replicate the first experiment on a real-world scene, detailed in \cref{sec:exp2}, and \cref{sec:exp3} presents an ablation study investigating the robustness of our two distance approaches to noise.
 Finally in \cref{sec:exp4} we validate our approach in a real-world remote teleoperation setup.

\subsection{Setup}\label{sec:setup}
\subsubsection{Scene reconstruction}\label{sec:evaluation:scene}
As described in \cref{sec:3DGS}, GS typically requires cene initialisation via SfM~\cite{schoenberger2016sfm}. Instead, we initialise GS directly from a subsampled RGB-D point cloud, preserving metric scale.
The input RGB-D point cloud is voxelised at 0.01\,m and uniformly sampled to 60k points. Normals are estimated to initialise the orientation of the 2D ellipsoids, accelerating training.
Unless otherwise stated, splats are trained for 1000 steps, taking under 40\,s on an NVIDIA RTX 3090 GPU.


\subsubsection{Robot platform}
All experiments are performed on a custom mobile manipulator platform, featuring a differential mobile base, Omron LD-60 and a 7-DoF Franka Panda arm mounted on top. The robot geometry is represented by a set of 77 manually placed spheres.

\subsubsection{Simulation details}
For all simulated experiments (\cref{sec:exp1,sec:exp2,sec:exp3}), the robot is simulated using the Swift simulator, part of the Robotics Toolbox in Python~\cite{Corke2021}.

\subsubsection{Computational  resources}
 All experiments were run on a desktop computer equipped with an Intel i7-12700K CPU and an NVIDIA RTX 3090 GPU. For \cref{sec:exp4}, the physical robot is controlled remotely via ROS.

\subsection{Experiment 1: Synthetic benchmark}\label{sec:exp1}

Our first set of experiments assesses the overall performance of the proposed controller in a reaching task in simulated synthetic scenes, with the goal of evaluating whether a controller operating directly on a Gaussian Splat (GS) representation can achieve performance comparable to methods that rely on perfect geometric information.

\subsubsection{Scenes}
\begin{figure}[t]
    \subfloat[Table]{
        \includegraphics[width=0.2\textwidth]{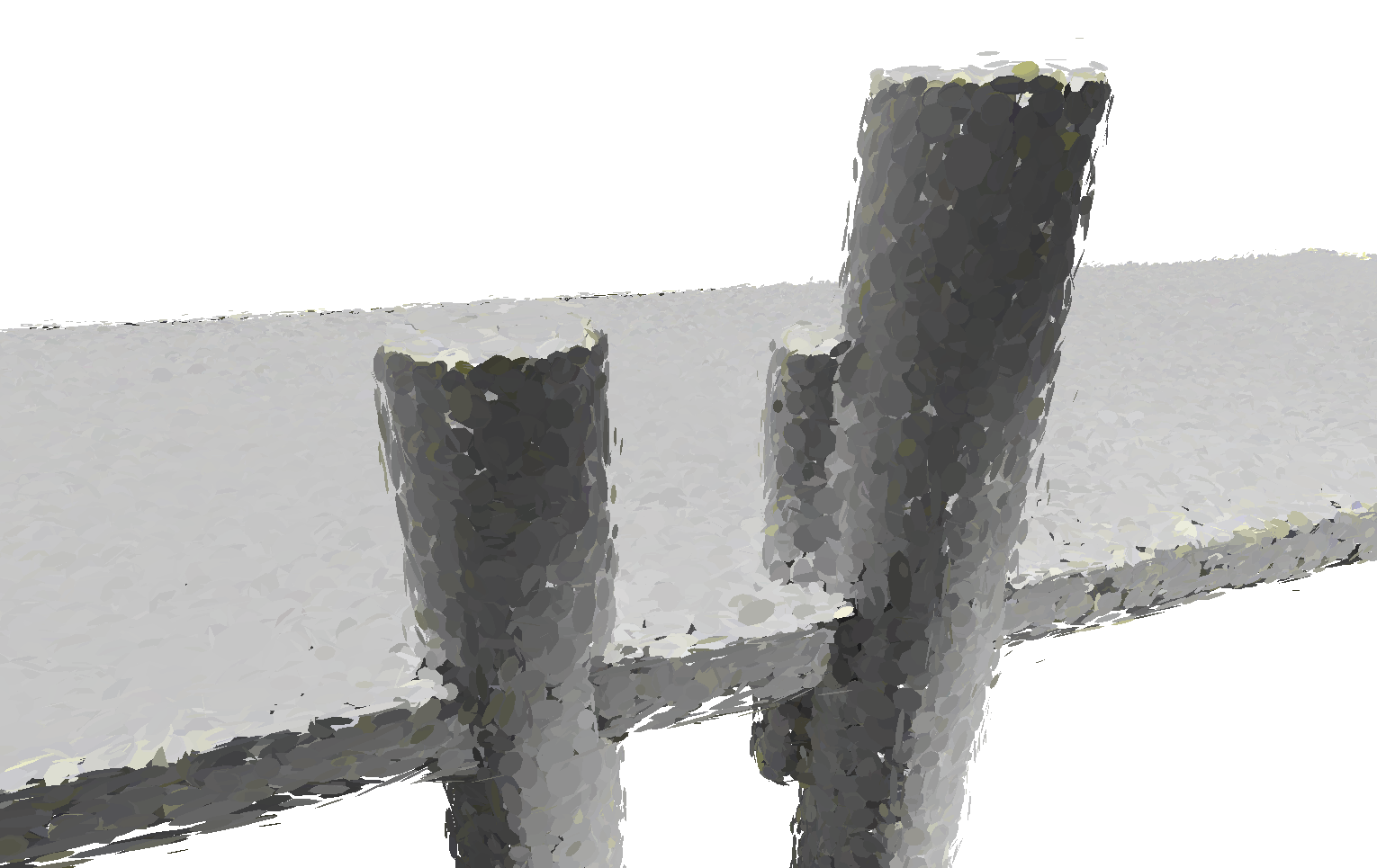}
    }
    \subfloat[Bookshelf]{
        \includegraphics[width=0.2\textwidth]{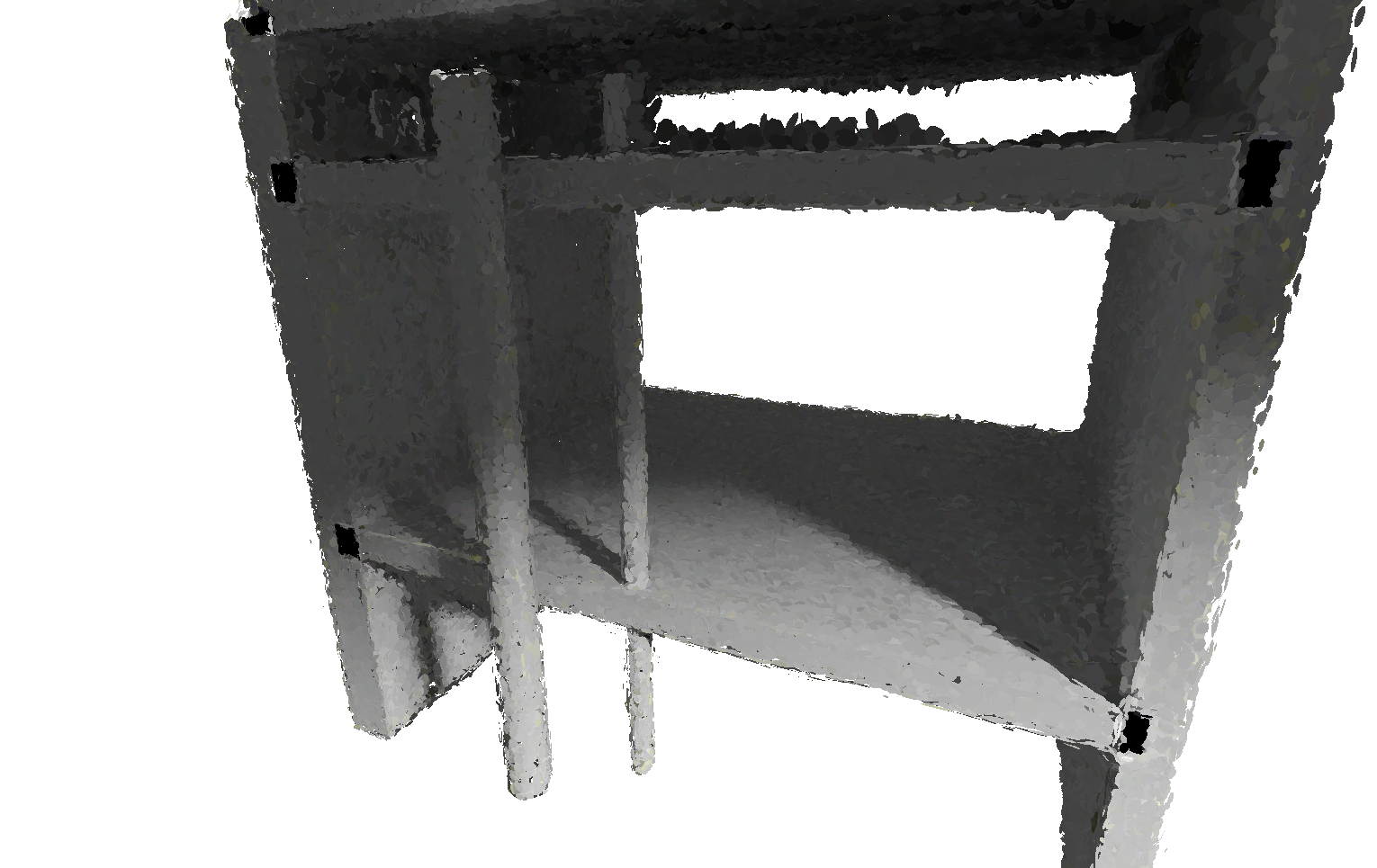}
    }
    \centering
    \caption{Resulting 2DGS Ellipsoids on a sample of the bookshelf and table scenarios utilised for our first set of experiments.}
    \label{fig:scenes}
    \vspace*{-0.3cm}
\end{figure}

The scenes used for this experiment are the ones proposed in~\cite{Marticorena2024RMMI:Map}, consisting of 500 variations of a clutter table and 500 variations of a bookshelf, in which reaching the target pose requires base motions. As these scenes only specify the location and scale of geometric primitives, we render RGB-D images in Blender and train a GS as described in \cref{sec:evaluation:scene}, with examples shown in \cref{fig:scenes}.

\subsubsection{Metrics}
The metrics used to assess the controller are:
\begin{itemize}
    \item \textbf{Success Rate:} The rate of the controller in generating a collision-free trajectory that reaches the desired 6-DoF pose with a margin of 2\,cm and 2$^\circ$.

    \item \textbf{Avg Distance:} 
    Mean minimum clearance from obstacles, computed using ground-truth distances $\hat{d}_j$ for each $j$-th sphere representing the robot:

    \begin{equation}
        D_{\text{Avg}} = \frac{1}{N_{\text{steps}}}\sum_{\tau = 0}^{N_{\text{steps}}}   \min_{j\in\mathcal{S}}{\hat{d}_j^{\tau}},
    \end{equation}
    where the superscript $\tau$ denotes each control step.

    \item \textbf{Gracefulness:} 
    Smoothness proxy measured as the average absolute end effector acceleration:
    \begin{equation}
        \left<\left| \vec{a}_{\text{eef}} \right|\right> = \frac{1}{N_{\text{steps}}}\sum_{\tau=0}^{N_{\text{steps}}} \left| \vec{a}^{\tau}_{\text{eef}} \right|.
    \end{equation}

    \item \textbf{End-effector path length:} Overall translation performed by the end effector during a successful trajectory.

    \begin{equation}
        L = \sum_{\tau=1}^{N_{\text{steps}}} \mid \vec{p}_{\text{eef}}^{\tau} -  \vec{p}_{\text{eef}}^{\tau - 1} \mid.
    \end{equation}

    \item \textbf{Collisions:} The rate of robot collisions with the environment.

    \item \textbf{QP Solving time:} Depending on the distance estimation approach, the number of collision constraints varies. This metric quantifies the overhead introduced to the overall optimisation.

\end{itemize}

\subsubsection{Baseline}
We compared our approach against~\cite{Marticorena2024RMMI:Map}, a Holistic QP controller that models collisions using a Signed Distance Field (SDF).

\subsubsection{Results}
\Cref{tab:main_table} reports the results, where metrics marked with \textbf{*} are computed only on trials in which all methods were successful. This helps eliminate bias toward artificially low obstacle distances or accelerations, which can occur when one variation becomes stuck in a local minimum and remains stationary near an obstacle for an extended period.
Our approach, which utilises a GS representation, achieves performance comparable to the baseline~\cite{Marticorena2024RMMI:Map} that relies on perfect distance information, validating that using GS does not deteriorate performance and highlighting the accuracy and reliability of the learned representation.

Furthermore, incorporating both soft and hard constraints into the control formulation leads to improved performance, as evidenced by an increase in success rates up to 8\%, safer motions by having a higher average distance to obstacles, with the cost of having 4\% longer end effector path lengths.
Notably, extending the formulation to include additional constraints by modelling distances along six distinct directions does not result in any significant increase in the overall solver; the QP solving times remain in the same order of magnitude, and importantly, well below the overall control rate.

In these experiments, scenes were represented using $60,813\pm246$ ellipsoids, yielding control frequencies of $17.2 \pm 1.3$\,Hz for the sphere-to-ellipsoid method and $24.2 \pm 2.7$\,Hz for the depth rasterisation approach.

\begin{table*}[tb]
\caption{Results of the synthetic scene benchmark.
Metrics are averaged over common successful trials (marked with \textbf{*}).
Best results are shown in \textbf{bold}, second best \underline{underlined}.
The proposed Gaussian Splatting–based controller achieves performance comparable to the ground-truth SDF baseline \cite{Marticorena2024RMMI:Map}.
Incorporating both hard and soft constraints improves success rates and clearance, while modelling distances in six directions adds negligible overhead to QP solving time.}
\label{tab:main_table}
\centering
\begin{tabular}{@{}cccccccccc@{}}
\toprule
Controller & Distance Method &  Inequalities & \begin{tabular}[c]{@{}c@{}}Active \\ Collision \\ Cost\end{tabular} & \begin{tabular}[c]{@{}c@{}}Success \\ Rate \\ (\%)$\uparrow$\end{tabular} & \begin{tabular}[c]{@{}c@{}}Avg Distance\textbf{*}\\ (m) $\uparrow$\end{tabular} & \begin{tabular}[c]{@{}c@{}}Gracefulness\textbf{*}\\ (m/s$^2) \downarrow$\end{tabular} & \begin{tabular}[c]{@{}c@{}}End effector\\ path length \textbf{*}\\ (m) $\downarrow$\end{tabular} & \begin{tabular}[c]{@{}c@{}}Coll.\\ (\%) $\downarrow$\end{tabular} & \begin{tabular}[c]{@{}c@{}}QP \\ solving time \\ (ms) $\downarrow$\end{tabular} \\ \midrule
\multirow{2}{*}{RMMI \cite{Marticorena2024RMMI:Map}} & \multirow{2}{*}{\begin{tabular}[c]{@{}c@{}}Ground Truth \\ global SDF\end{tabular}} & \multirow{2}{*}{$|\mathcal{R}|$} & $\times$ & 0.778 & 0.087 & 0.388 & \underline{0.724} & \textbf{0} & \textbf{0.15} $\pm$ \textbf{0.04} \\
 &  &  & $\checkmark$ & \underline{0.846} & \underline{0.112} & \underline{0.279} & 0.759 & \textbf{0} & \textbf{0.15} $\pm$ \textbf{0.04} \\ \midrule
\multirow{4}{*}{Ours} & \multirow{2}{*}{\begin{tabular}[c]{@{}c@{}}Sphere-to-ellipsoid\\ \cref{sec:euclidean}\end{tabular}} & \multirow{2}{*}{$|\mathcal{R}|$} & $\times$ & 0.776 & 0.087 & 0.434 & \textbf{0.723} & \textbf{0} & $0.17 \pm 0.08$ \\
 &  &  & $\checkmark$ & \textbf{0.856} & \textbf{0.113} & 0.291 & 0.762 & \textbf{0} & \underline{$0.17 \pm 0.03$} \\ \cline{2-10} 
 & \multirow{2}{*}{\begin{tabular}[c]{@{}c@{}}Depth rasterisation\\ \cref{sec:distance_depth}\end{tabular}} & \multirow{2}{*}{$|\mathcal{R}| \times 6$} & $\times$ & 0.8 & 0.086 & 0.411 & 0.726 & \textbf{0} & $0.21 \pm 0.06$ \\
 &  &  & $\checkmark$ & 0.845 & 0.107 & \textbf{0.272} & 0.754 & \textbf{0} & $0.20 \pm 0.06$ \\ \bottomrule
\end{tabular}
\end{table*}

\subsection{Experiment 2: Qualitative Real-world scene}
\label{sec:exp2}

\begin{figure}[t]
    \vspace{0.1cm}
    \centering
    \subfloat[Clean reconstruction]{
        \includegraphics[width=0.7\linewidth,trim={0 3cm 6cm 7cm},clip]{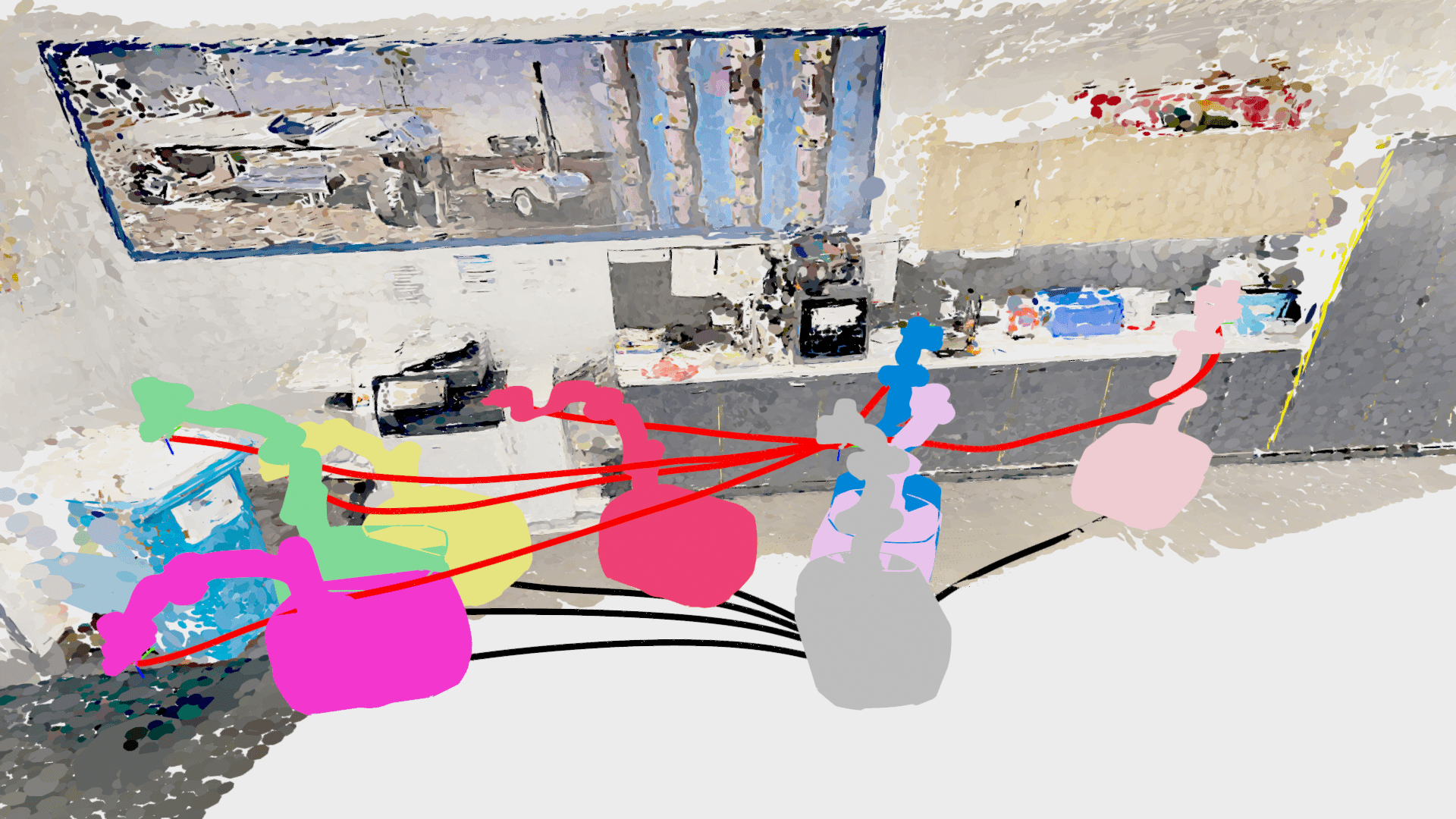}
        \label{fig:scene:clean}
    }

          
    \subfloat[Noisy reconstruction]{
        \includegraphics[width=0.7\linewidth,trim={0 3cm 6cm 7cm},clip]{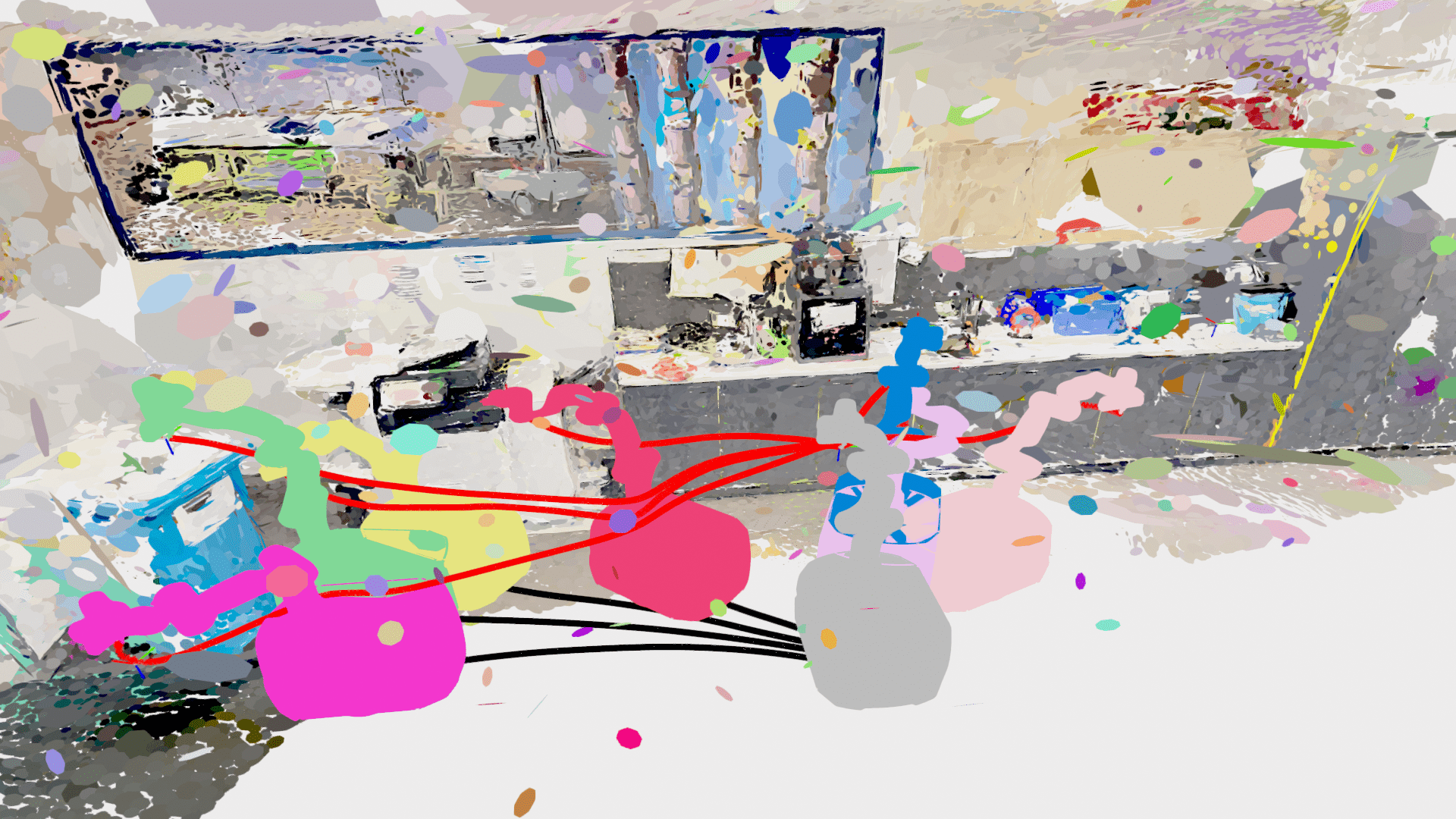}\label{fig:scene:noisy}
    }

    \caption[ 2DGS reconstruction results on real-world scene]{Resulting Gaussian Splats of the printer scene, including the resulting motions showing in red and black the trajectories of the end effector and the robot base, respectively, on (a) Clean reconstruction of the scene in which the ellipsoids align with the underlying geometry. (b) Noisy reconstruction obtained by adding 1,000 random points before training, in which we can see a set of floating ellipsoids.}
    \label{fig:scene}
\end{figure}
We further demonstrate our method using a real-world scan of an office space. To do so, the scene was captured offline by aggregating 30 posed RGB and depth images obtained using an iPhone 14 Pro. We obtain the poses of each frame using \textit{ARKit} (\url{https://developer.apple.com/augmented-reality/arkit/}) through the NeRF Capture app~\cite{NeRFCapture} to maintain a representation in metric scale. The reconstructed scene is presented in \cref{fig:scene:clean}.

Due to the lack of an accurate ground truth geometry of the scene, we only compare the average acceleration and the success rate of reaching the desired end effector pose while not violating the constraints (i.e., colliding). We manually place seven targets for the robot to reach, including one target placed behind a cupboard to test the controller when the pose is unreachable.

The resulting metrics of these simulated trajectories are presented in \cref{tab:real_world},
and a qualitative depiction of the resulting motions is shown in \cref{fig:scene:clean}.
As we can observe in a scenario where the ellipsoids align with the surface, we qualitatively obtain the same motions.

\begin{table}[tb]
\centering
\caption[Real-world scan experiments]{Results of the real-world scan experiment. Both distance methods achieved similar success rates, with no collisions.}
\label{tab:real_world}
\begin{tabular}{@{}ccccc@{}}
\toprule
Distance Method &
  Success Rate $\uparrow$ &
  \begin{tabular}[c]{@{}c@{}}Gracefulness\textbf{*}\\ $\left \langle |\vec{a}| \right \rangle (m/s^2) \downarrow$\end{tabular} &
  \multicolumn{1}{l}{Collisions} \\ \midrule
\begin{tabular}[c]{@{}c@{}}Sphere-to-ellipsoid\\ \cref{sec:euclidean}\end{tabular} &
  6/7 &
  0.148 &
  0 \\
\begin{tabular}[c]{@{}c@{}}Depth rasterisation\\ \cref{sec:distance_depth}\end{tabular} &
  6/7 &
  0.196 &
  0 \\ \bottomrule
\end{tabular}
\vspace*{-0.3cm}
\end{table}

\subsection{Experiment 3: Performance under noisy reconstructions}\label{sec:exp3}
Our third experiment evaluates both distance methods in scenes with noisy floaters.
We explore noisy scenarios by adding 1,000 random points to the 50,000 points used to initialise the real-world captured scene; these added points aim to simulate a scenario with poor scene recording. We proceed to train the splats using this set of 51,000 points; the resulting ellipsoids are shown in \cref{fig:scene:noisy}.

We tested both distance methods on this suboptimal reconstruction. The metrics are presented in \cref{tab:real_world_noise}, and a depiction of the motions in \cref{fig:scene:noisy}. We observe that the sphere-to-ellipsoid approach fails catastrophically by considering false positive collisions due to not handling the opacities of each ellipsoid. In contrast, the depth rasterisation effectively filters low-opacity ellipsoids by considering the overall transmittance of all virtual cameras. However, it can still get stuck (but not collide) in some regions due to false positive predictions.

The results of this experiment highlight the importance of obtaining a high-quality reconstruction. In this experiment, we purposely added noise, highlighting that our sphere-to-ellipsoid distance approach is more susceptible to noise compared to a depth rasterisation approach.

\begin{table}[tb]
\centering
\caption[Real world scans]{Results with noisy scene reconstructions. Noise degrades performance significantly for the sphere-to-ellipsoid method.}
\label{tab:real_world_noise}
\begin{tabular}{@{}cccc@{}}
\toprule
Distance Method &
  Success Rate $\uparrow$ &
  \begin{tabular}[c]{@{}c@{}}Gracefulness\textbf{*}\\ $\left \langle |\vec{a}| \right \rangle (m/s^2) \downarrow$\end{tabular} &
  \multicolumn{1}{l}{Collisions} \\ \midrule
\begin{tabular}[c]{@{}c@{}}Sphere-to-ellipsoid\\ \cref{sec:euclidean}\end{tabular} &
  0/7 &
  N/A &
  N/A \\
\begin{tabular}[c]{@{}c@{}}Depth rasterisation\\ \cref{sec:distance_depth}\end{tabular} &
  4/7 &
  0.281 &
  0 \\ \bottomrule
\end{tabular}
\vspace*{-0.3cm}
\end{table}
   
\subsection{Experiment 4: Real robot teleoperation}
\label{sec:exp4}

Our final experiment aims to validate one of the use cases of our approach, namely, aided teleoperation. The goal of this experiment is to validate the reactivity of our approach to track a target end-effector pose input from a teleoperator, while maintaining a safe distance within the scene.

For these experiments, we warm-start the scene representation by collecting 8 posed RGB-D frames from a RealSense D435 camera mounted on the end-effector. These posed RGB-D frames are used to train an initial set of splats. Once the splats are trained, we provide the teleoperator a 3D visual of the current robot state alongside the current splats using an open-source 3D viewer that incorporates a GS viewer \cite{yi2025viser}. 
From this interface, the teleoperator can control the target end-effector pose in real time while the training is paused. From this interface, the teleoperator can trigger updates to the GS representation by collecting additional frames from the robot or by refining the model by training splats on the new frames for additional steps. We refer the reader to the accompanying video for additional details.

\begin{figure}[t]
\centering
\includegraphics[width=\linewidth]{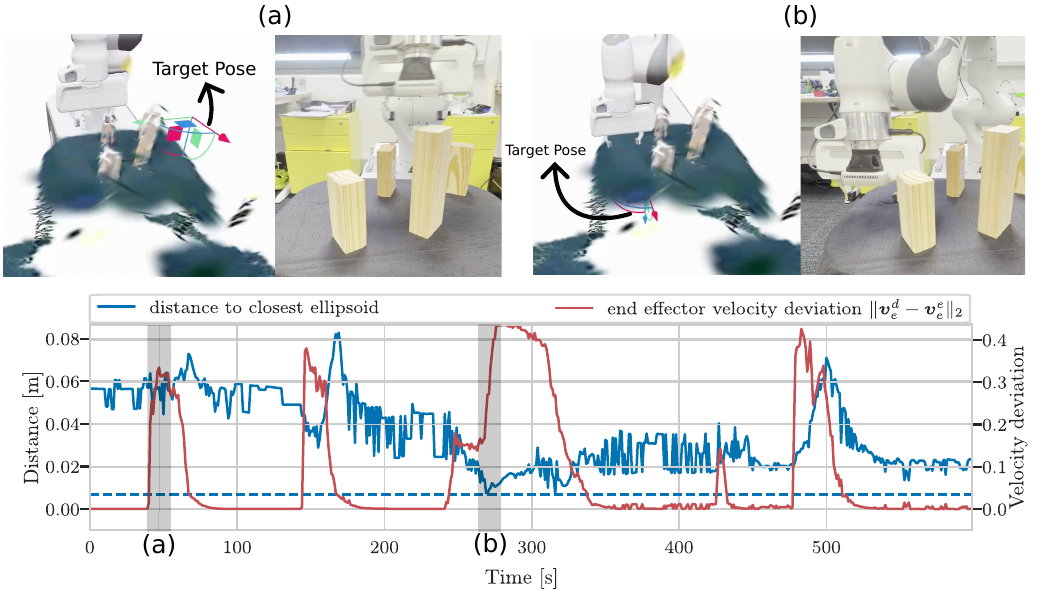}
\centering
\vspace*{-0.5cm}
\caption{Snapshots of the resulting motion while being teleoperated. 
(Top) We show a side-by-side view of the teleoperation GUI, displaying the current robot state and the highlighted target pose, accompanied by an image of the real robot.
(Bottom) We show the distance between the robot and the scene, alongside the deviation between the commanded velocity and the velocity executed by the controller. As we can see, the controller effectively avoids collisions, even in scenarios such as the one highlighted in (b), where the target end-effector pose is not reachable.}
\label{fig:RealWorld}
\vspace*{-0.3cm}
\end{figure}

Additionally, \cref{fig:RealWorld} shows a trajectory in which the teleoperator forcefully attempts to collide with the scene; snapshots of this trajectory are accompanied by the robot's distance to the scene $\tilde{d}(t)=\min_{j\in\mathcal{S}}(d_j(t))$, alongside the difference between the desired end-effector velocity $\vec{v}_{e}^{d}$ (computed as straight line towards the target pose) and the executed end-effector velocity $\vec{v}_{e}^{e}$ derived from our controller. We observe that the controller deviates from a straight-line path to maintain a safe distance to the scene, even in scenarios where the target end-effector pose is non-reachable.


\section{Conclusions}
In this paper, we proposed a mobile manipulation controller that utilises the high-fidelity reconstruction of a Gaussian Splat. 
Our results in synthetic scenes suggest that, in an idealised scenario, both depth rasterisation and ellipsoid distance can be used to extract the surface of the captured scene, informing our controller about occupied space.
Further experiments validate that in the presence of low-opacity ellipsoids, the depth rasterisation approach provides more robust results by effectively filtering them.
We further validated the approach on a real robot platform, demonstrating that the controller can track target end-effector poses while maintaining safe distances to the environment.

Our current evaluation focuses on static scenes. Handling structural changes in the environment could be achieved by integrating recent change-detection and update mechanisms for GS models~\cite{change_detection_gs, 3dgs_changes}.
More broadly, since our controller directly operates on a GS representation, it opens several promising research directions, such as integrating dynamic gaussians \cite{luiten2023dynamic, abou-chakra2024physically} or semantics-aware splats~\cite{Qin2024}, enabling task-aware interaction with complex environments.


\bibliography{references}
\bibliographystyle{IEEEtran} 

\end{document}